\documentclass[letterpaper, 10 pt, conference]{ieeeconf}  

\IEEEoverridecommandlockouts                              

\usepackage{booktabs}       
\usepackage{amsfonts}       
\usepackage{amsmath,amssymb}
\usepackage{graphicx}
\usepackage{multirow}
\usepackage[export]{adjustbox}
\usepackage{xcolor}
\usepackage{subcaption}
\usepackage{tabularx}
\usepackage{siunitx}
\usepackage{csquotes}

\allowdisplaybreaks

\makeatletter
\let\NAT@parse\undefined
\makeatother

\usepackage[breaklinks=true,colorlinks,bookmarks=false]{hyperref}
\usepackage[capitalise]{cleveref}

\usepackage{url}

\renewcommand{\baselinestretch}{0.97}

\title{\LARGE \bf
The Impact of Class Uncertainty Propagation in \\  Perception-Based Motion Planning
}

\author{Jibran Iqbal Shah$^{1}$, Andrei Ivanovic$^{1}$, Kelly Zhu$^{1}$, Masha Itkina$^{2}$,\\
        Rowan McAllister$^{2}$, Igor Gilitschenski$^{1}$, Florian Shkurti$^{1}$%
\thanks{$^{1}$Department of Computer Science, University of Toronto. \newline
        {\tt\footnotesize \{a.ivanovic, jibraniqbal.shah\}@mail.utoronto.ca, \{zhu, gilitschenski, florian\}@cs.toronto.edu}}%
\thanks{$^{2}$Toyota Research Institute. \newline
        {\tt\footnotesize \{masha.itkina, rowan.mcallister\}@tri.global}}%
}%

\begin{document}

\maketitle

\thispagestyle{empty}
\pagestyle{empty}


\begin{abstract}
Autonomous vehicles (AVs) are being increasingly deployed in urban environments. In order to operate safely and reliably, AVs need to account for the inherent uncertainty associated with perceiving the world through sensor data and incorporate that into their decision-making process. Uncertainty-aware planners have recently been developed to account for upstream perception and prediction uncertainty. However, such planners may be sensitive to prediction uncertainty miscalibration, the magnitude of which has not yet been characterized. Towards this end, we perform a detailed analysis on the impact that perceptual uncertainty propagation and calibration has on perception-based motion planning. We do so by comparing two novel prediction-planning pipelines with varying levels of uncertainty propagation on the recently-released nuPlan planning benchmark. We study the impact of upstream uncertainty calibration using closed-loop evaluation on the nuPlan challenge scenarios. We find that the method incorporating upstream uncertainty propagation demonstrates superior generalization to complex closed-loop scenarios.
\end{abstract}

\section{INTRODUCTION}
Autonomous vehicles (AVs) are becoming increasingly adopted and deployed in highly-dynamic and uncertain urban environments. Uncertainty is present almost everywhere AVs perceive a scene, whether it arises from multimodality in an agent’s potential future motion or measurement uncertainty from onboard sensors and occlusions. Despite this, most AVs operate without capturing this inherent uncertainty, even with some modules providing highly uncertain outputs~\cite{IvanovicLeeEtAl2022, HenneSchwaiger2019}. To improve their safety and robustness, AVs should adopt autonomy stacks that are uncertainty-aware and can use such information to better inform AV motion planning.

Currently, AV autonomy stacks are commonly architected with four main components: perception, prediction, planning, and control. While most trajectory predictors do not incorporate upstream sources of uncertainty, several recent works have augmented state-of-the-art (SOTA) methods to incorporate and implicitly propagate upstream perceptual uncertainty, such as state or class uncertainty~\cite{IvanovicLeeEtAl2022, IvanovicLinEtAl2022}. In doing so, these methods have shown improved results, both in traditional prediction metrics (e.g., displacement errors) and in model calibration; how closely a model's predicted uncertainty aligns with its empirical uncertainty, which can be defined as the fraction of ground truth futures that lie within specified probability thresholds~\cite{ovadia2019can, IvanovicHarrison2023}. 

\begin{figure}[ht]
    \centering
        \begin{flushright}
    \includegraphics[width=0.9\linewidth]{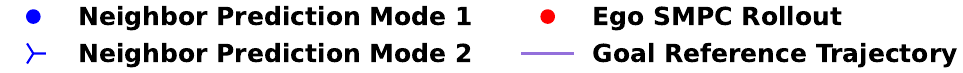} \\  
    \end{flushright}
    \vspace{-0.5em}
    \begin{tabular}{l}  
        \includegraphics[width=0.982\linewidth]{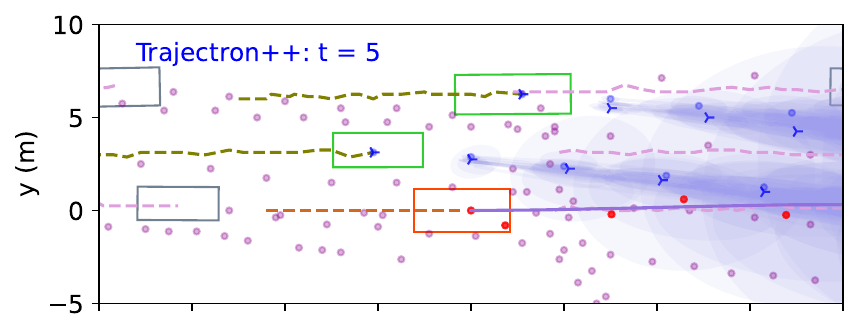} \\[-0.5em]
        \includegraphics[width=\linewidth]{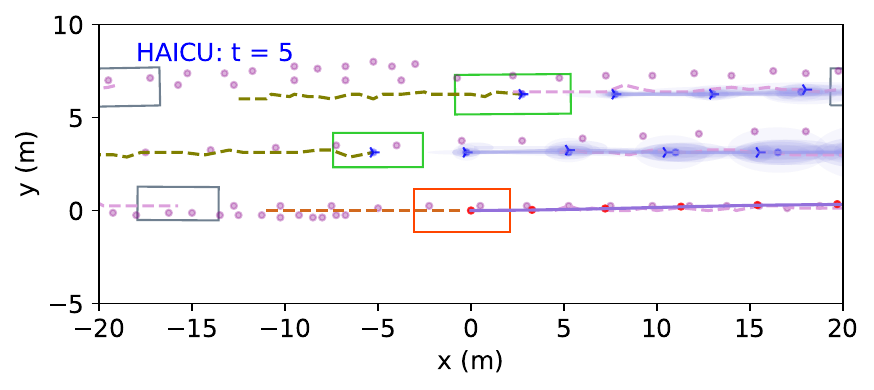} \\[-0.5em]
    \end{tabular}
    \vspace{-0.5em}  
    \caption{A direct comparison of prediction-planning performance between our novel uncertainty-aware stochastic MPC (UA-SMPC) when paired with Trajectron++~\cite{SalzmannIvanovicEtAl2020}  (left), an uncertainty-agnostic prediction model, and HAICU~\cite{IvanovicLeeEtAl2022} (right), an uncertainty-aware prediction model. HAICU produces significantly more calibrated trajectory predictions than Trajectron++. As a result, HAICU-backed UA-SMPC generates smoother, lane-respecting motion plans, while Trajectron++-backed UA-SMPC swerves aggressively to avoid collisions with the poorly-calibrated predictions. Ego vehicle is shown in orange, nearby non-ego vehicles in green, and distant non-ego vehicles in grey with pink tracks.}
    \label{fig:teaser}
      \vspace{-1.5em}
\end{figure}

In parallel, uncertainty-aware planners have recently been developed that account for prediction uncertainty and generate motion plans which are more efficient without sacrificing safety~\cite{NairEtAl2022, NairTsengBorelli2022, NairTsengBorelliCDC2022}. However, such planners may be sensitive to prediction uncertainty miscalibration, manifested for example as improperly-shaped covariances in a predictor's output distribution. While there has been work comparing the impact of class uncertainty propagation on such planners in the open-loop setting~\cite{IvanovicItkinaEtAl2024}, open-loop evaluation resets the ego state to ground truth at each timestep, so the downstream effects of planning decisions cannot propagate to future steps, masking the compounding effects of prediction miscalibration that this work aims to characterize. Furthermore, the magnitude of this sensitivity remains unknown, as there have been no sensitivity analysis studies that measure the effect of prediction uncertainty calibration on uncertainty-aware, perception-based motion planning. Towards this end, we seek to determine the importance of perception and prediction output uncertainty calibration on motion planning.

{\bf Contributions.} Our contributions are threefold:
\begin{itemize}
    \item A novel uncertainty-aware stochastic MPC (UA-SMPC) formulation extending~\cite{NairEtAl2022} that directly incorporates multi-modal GMM predictions from multiple agents with an exact probabilistic collision avoidance constraint.
    \item A detailed analysis of the impact of class uncertainty propagation on perception-based motion planning in the closed-loop setting, evaluated on nuPlan's large-scale planning benchmark~\cite{caesar2021nuplan}. We compare two SOTA trajectory forecasting methods, HAICU~\cite{IvanovicLeeEtAl2022} (uncertainty-aware, with upstream uncertainty propagation) and Trajectron++~\cite{SalzmannIvanovicEtAl2020} (uncertainty-agnostic), both paired with our UA-SMPC planner.
    \item A sensitivity analysis of our planning framework to prediction covariance calibration, characterizing how scaling the predicted covariances affects closed-loop planning performance.
\end{itemize}

\section{RELATED WORK}\label{sec:litreview}

{\bf Autonomous Vehicle Trajectory Forecasting.} Recent trajectory forecasting methods employ Conditional Variational Autoencoders (CVAEs)~\cite{SohnLeeEtAl2015} as in~\cite{RhinehartMcAllisterEtAl2019, IvanovicHarrison2023}, Graph Neural Networks (GNNs)~\cite{ScarselliGoriEtAl2009} as in~\cite{lanegcn}, and Transformers~\cite{VaswaniShazeerEtAl2017} as in~\cite{huang2023gameformer}, ~\cite{mmtransformer} which have been used to explicitly model and capture all possible agent interactions and future trajectories in a scene~\cite{RudenkoPalmieriEtAl2019, HuangEtAl2022}.

\begin{figure}[h]
    \centering
    \includegraphics[width=0.96\linewidth]{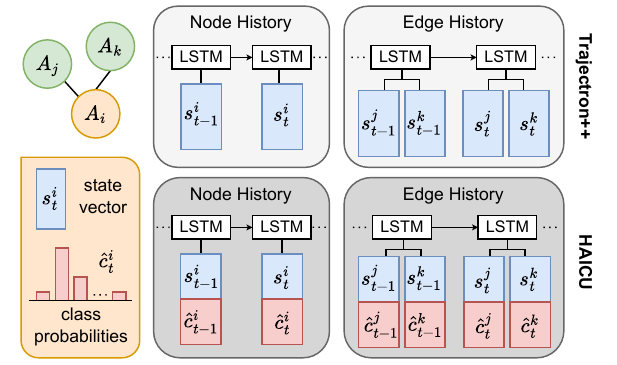}
    \caption{In both Trajectron++ and HAICU, the scene is modeled as a graph with nodes representing agents and edges representing their interactions. However, one key distinction is that HAICU incorporates both the state vector and class probabilities into its node and edge histories, whereas Trajectron++ only includes the state vector.}
    \label{fig:architecture}
\end{figure}

In this work, we employ Trajectron++~\cite{SalzmannIvanovicEtAl2020} and its follow-up work HAICU~\cite{IvanovicLeeEtAl2022}. Both models use the exact same Trajectron++ backbone, producing Gaussian Mixture Model (GMM) predictions. HAICU additionally propagates class uncertainty, enabling direct comparison of the isolated effect of uncertainty on downstream planning. ~\cref{fig:architecture} further elaborates on the key architectural difference between Trajectron++ and HAICU. Moreover, PSU-TF~\cite{IvanovicLinEtAl2022} extends Trajectron++~\cite{SalzmannIvanovicEtAl2020} to account for state uncertainties stemming from detection and tracking. A detailed survey of uncertainty estimation and quantification methods can be found in~\cite{gawlikowski2021survey}. Each of those works perform uncertainty estimation, but as we are dealing with a modular autonomous driving stack, each component is able to characterize its own uncertainty and provide it to following modules. Both Trajectron++ and HAICU are well-established, achieving SOTA performance when released, with easy-to-use publicly-available codebases.

{\bf Stochastic MPC for Autonomous Driving.} For autonomous driving scenarios with a high density of overlapping, non-Markovian, multi-modal predictions, many SMPC works exploit the structure of GMMs~\cite{zhou2018joint, wang2020non, ren2022chance, NairEtAl2022, zhou2023interaction} due to their memory-efficiency in representing multi-modal uncertainty~\cite{nair2023predictive}. To address the conservatism of emerging non-linear, non-convex SMPC methods~\cite{zhou2018joint, wang2020non}, a recent work~\cite{NairEtAl2022} proposes a convex formulation optimizing over a novel class of policies, enhancing the feasibility of the optimization problem. In this work, we extend their formulation to directly incorporate uncertainty, multiple agents, and more sophisticated ego dynamics.

\section{Input Uncertainty-Aware Stochastic MPC}\label{sec:method}

To investigate the effects of perception and prediction output uncertainty calibration on motion planning, we provide the predictions of Trajectron++ and HAICU to a novel uncertainty-aware stochastic MPC (UA-SMPC) formulation. Our formulation extends the SMPC framework presented in~\cite{NairEtAl2022}. In practice, we found that the convex over-approximations of collision avoidance regions in~\cite{NairEtAl2022} become overly restrictive when multiple agents are present, leading to infeasibility or excessively conservative behavior. To address this, we devise a collision avoidance constraint that avoids the minimal keep-out region required to guarantee $\Pr(\text{no collision}) \geq p$ for a chosen per-step coverage level $p$, neither over-approximating nor under-approximating this region. This property is particularly important for our study, which measures the impact of miscalibrated prediction uncertainties on motion planning; we require that any observed effects stem from the prediction covariances themselves, not from artifacts introduced by constraint approximations. While our formulation results in a non-convex optimization problem, we solve it using CasADi~\cite{Andersson2019} with the IPOPT solver and find it to be feasible in practice. We also expand upon~\cite{NairEtAl2022} to incorporate recent trajectory forecasting approaches that provide multi-modal agent predictions. For consistency, we use similar notation as presented in~\cite{NairEtAl2022}.

\subsection{Notation}
\label{sec:smpc_notation}

Let $x_t \in \mathbb{R}^2$ represent the 2D position of the ego-vehicle (EV) at time $t$, with $x_t = [X_t, Y_t]^\top$. The control inputs $u_t = [(u_t)_x, (u_t)_y]^\top$ are the vehicle's longitudinal and lateral velocities. For all non-ego vehicles (NEVs), $o_t^i \in \mathbb{R}^2$ denotes the 2D position of NEV $i \in \{1, \dots, A\}$, where $A$ is the total number of neighboring agents at time $t$. We denote the future $N$-step predictions of the $i$-th NEV's position as random variables $\{o_{k|t}^i\}_{k=1}^N$. As per~\cite{NairEtAl2022}, $o_{k|t}^i$ is a shortened notation for $o_{t+k}^i|o_{t}^i$. The output predictions for a single NEV at time $t$ are encoded as a bivariate GMM, $\mathcal{G}_t^i = \left\{ p_{t}^{i,j}, \{\mathcal{N}(\mu_{k|t}^{i,j}, \Sigma_{k|t}^{i,j})\}_{k=1}^N \right\}_{j=1}^\Xi$, where $\Xi$ is the number of considered modes and $p_{t}^{i,j} = \mathbb{P}(\sigma_t^i = j \mid o_t^i)$ denotes the probability of the $j$-th mode at time $t$. Following~\cite{NairEtAl2022}, we use a shared mode index $j$ across all agents rather than independent per-agent modes, avoiding the combinatorial explosion of $\Xi^A$ joint modes. We impose the collision avoidance constraint for each mode $j$ at each timestep; since the constraint guarantees $\Pr(\text{no collision} \mid \text{mode } j) \geq p$ for every mode, the marginal probability $\Pr(\text{no collision}) \geq p$ follows by the law of total probability.

\subsection{Overview of SMPC Formulation}

The SMPC formulation in this work is used by the EV to track a reference trajectory safely and comfortably, while using multi-modal predictions for collision avoidance and navigating interactions with other agents. We employ deterministic NEV predictions by setting $f_k^{\text{NEV}}(o_{0|t}^{i,j}) = \hat{o}_{k|t}^{i,j} = \mu_{k|t}^{i,j}$, i.e., the mean of the $j$-th GMM component for NEV $i$. The optimization problem of our SMPC framework is formulated as
\begin{align}
    \min_{h^j, K^j} \quad & J_t(\mathbf{x}_t, \mathbf{u}_t), \label{eq:smpc_obj}\\
    \text{s.t.} \quad & x_{k+1|t}^j = f_k^\text{EV}(x_{k|t}^j, u_{k|t}^j), \label{eq:smpc_dynamics}\\
    & v_{\min} \leq u_{k|t}^j \leq v_{\max}, \label{eq:smpc_vel}\\
    & a_{\min} \leq u_{k+1|t}^j - u_{k|t}^j \leq a_{\max}, \label{eq:smpc_accel}\\
    & \text{dist}_2\!\left(\Sigma_{k+1|t}^{i,j^{-1/2}}(x_{k+1|t}^j - \mu_{k+1|t}^{i,j}),\, \mathcal{B}^{i,j}_{k+1|t}\right) \geq \sqrt{\beta}, \label{eq:smpc_collision}\\
    & x_{0|t} = x_t,\ o_{0|t}^i = o_t^i, \label{eq:smpc_init}\\
    & \forall\ k \in \{0, \dots, N-1\}, \notag \\ 
    & \forall\ i \in \{1, \dots, A\}, \notag \\
    & \forall\ j \in \{1, \dots, \Xi\}, \notag
\end{align}
where $\mathbf{x}_t = [x_{0|t}, \dots, x_{N-1|t}]$ and $\mathbf{u}_t = [u_{0|t}, \dots, u_{N-1|t}]$.

The objective function~\eqref{eq:smpc_obj} penalizes the EV's deviation from a kinematically-feasible reference trajectory. Constraints~\eqref{eq:smpc_vel} and~\eqref{eq:smpc_accel} enforce velocity and acceleration bounds, respectively. The collision avoidance constraint~\eqref{eq:smpc_collision} ensures probabilistic safety by requiring the EV to maintain sufficient distance from the uncertainty-adjusted keep-out region $\mathcal{B}^{i,j}_{k+1|t}$, which is the overlap rectangle $\mathcal{R}$ (representing vehicle geometries) transformed into Mahalanobis coordinates using the predicted covariance $\Sigma_{k+1|t}^{i,j}$; the full derivation is given in~\cref{sec:collision_avoidance}. EV and NEV state feedback are incorporated via the initial conditions in~\eqref{eq:smpc_init}. The different modes are \enquote{coupled} together by having a shared first control $h_{0|t}^j = h_0$ for all $j$, following a similar formulation in~\cite{NairEtAl2022}.

\subsection{Ego-Vehicle Dynamics Model}

For the EV dynamics $f_k^\text{EV}(x_{k|t}^j, u_{k|t}^j)$, we follow an affine-linear (in velocity) dynamics model where the next state $x_{k+1|t}^j = [X_{k+1|t}^j, Y_{k+1|t}^j]^\top$ is given by
\begin{align}
    X_{k+1|t}^j &= X_{k|t}^j + \delta_t (u_{k|t}^j)_x, \label{eq:ev_dynamics_x}\\
    Y_{k+1|t}^j &= Y_{k|t}^j + \delta_t (u_{k|t}^j)_y, \label{eq:ev_dynamics_y}
\end{align}
where $\delta_t$ is the discretization timestep.

\subsection{Collision Avoidance}
\label{sec:collision_avoidance}

We formulate collision avoidance by combining rectangular vehicle geometries with ellipsoidal uncertainty regions derived from the GMM predictions. Both the EV and NEV are modeled as axis-aligned rectangles with half-lengths $R_1$ (longitudinal) and $R_2$ (lateral). Two such rectangles collide if and only if their centers are within distance $R_1$ longitudinally and $R_2$ laterally. This defines the center-to-center overlap region as the rectangle
\begin{equation}
    \mathcal{R} := \{r \in \mathbb{R}^2 : |r_1| \leq R_1,\ |r_2| \leq R_2\},
\end{equation}
which is the Minkowski sum of the two half-size rectangles. Collision occurs if and only if $x - o \in \mathcal{R}$, where $x$ is the EV center and $o$ is the NEV center. Under our GMM predictions, the NEV center $o$ is a random variable. For a given mode $j$, we have $o \sim \mathcal{N}(\mu, \Sigma)$ where $\mu = \mu_{k|t}^{i,j}$ and $\Sigma = \Sigma_{k|t}^{i,j}$. For a per-step coverage level $p \in (0, 1)$, let $\beta = \chi^2_{2,p}$ be the chi-square quantile with 2 degrees of freedom. The Mahalanobis ellipsoid
\begin{equation}
    E := \{y \in \mathbb{R}^2 : (y - \mu)^\top \Sigma^{-1} (y - \mu) \leq \beta\}
\end{equation}
contains the NEV center with probability $p$. Since the NEV center $o$ could lie anywhere in $E$, a collision is possible for some realization $o \in E$ if and only if $x \in E \oplus \mathcal{R}$, where $\oplus$ denotes the Minkowski sum. The keep-out requirement $x \notin E \oplus \mathcal{R}$ guarantees collision is avoided for all $o \in E$; since $\Pr(o \in E) = p$, this implies $\Pr(\text{no collision}) \geq p$. Crucially, this is the minimal region that achieves this guarantee as any smaller keep-out region would violate the probability bound. To evaluate this constraint efficiently, we transform to Mahalanobis coordinates $z := \Sigma^{-1/2}(x - \mu)$. Under this map, the ellipsoid $E$ becomes the disk $\mathbb{B}_2(\sqrt{\beta}) := \{z : \|z\|_2 \leq \sqrt{\beta}\}$, and the overlap rectangle $\mathcal{R}$ maps to the parallelogram (zonotope)
\begin{equation}
    \mathcal{B} := \Sigma^{-1/2} \mathcal{R} = \{s_1 v_1 + s_2 v_2 : |s_1|, |s_2| \leq 1\},
\end{equation}
where $v_1 = \Sigma^{-1/2} [R_1, 0]^\top$ and $v_2 = \Sigma^{-1/2} [0, R_2]^\top$. Since Minkowski summing with a Euclidean ball is exactly a distance dilation, the exact, tight collision avoidance check becomes
\begin{equation}
\label{eq:collision_constraint}
    \text{dist}_2\!\left(\Sigma^{-1/2}(x - \mu),\, \Sigma^{-1/2}\mathcal{R}\right) \geq \sqrt{\beta},
\end{equation}
where $\text{dist}_2(z, \mathcal{B}) = \min_{s \in \mathcal{B}} \|z - s\|_2$ is the Euclidean distance from $z$ to the set $\mathcal{B}$. This can be computed in closed form: with $V = [v_1\ v_2] \in \mathbb{R}^{2 \times 2}$, we have $\text{dist}_2(z, \mathcal{B}) = \min_{|s_1|, |s_2| \leq 1} \|Vs - z\|_2$. Since $\mathcal{B}$ is a convex polygon in 2D, the nearest point is found by checking the projection onto the interior, the four edges, and the four corners. This constraint is applied for each NEV $i$, each mode $j$, and each timestep $k$ in~\eqref{eq:smpc_collision}, where $\mathcal{B}^{i,j}_{k+1|t} = \Sigma_{k+1|t}^{i,j^{-1/2}} \mathcal{R}$.

\subsection{Parameterized Policies Incorporating State Feedback}

We optimize over a class of parameterized policies that incorporate NEV state feedback, following a similar structure to~\cite{NairEtAl2022}. Note that all positions are expressed relative to the EV position at the current timestep, as provided by our prediction model. The feedback policy is designed such that the first control is shared across all modes while subsequent controls are mode-specific, enabling the policy to adapt to the realized NEV behavior. For $k = 0$, the control is shared across all modes and uses the actual observed NEV positions:
\begin{equation}
    u_{0|t} = h_{0|t} + \sum_{i=1}^A K^i o_t^i.
\end{equation}
For $k \geq 1$ and for each mode $j \in \{1, \dots, \Xi\}$:
\begin{equation}
    u_{k|t}^{j} = h_{k|t}^j + \sum_{i=1}^A K_{k|t}^{i,j} \hat{o}^{i,j}_{k|t},
\end{equation}
where $\hat{o}^{i,j}_{k|t} = \mu_{k|t}^{i,j}$ is the predicted mean position of NEV $i$ under mode $j$. This structure couples the modes at the first timestep through the shared feedforward term $h_{0|t}$ and feedback gains $K^i$, while allowing mode-specific behavior for subsequent timesteps through $h_{k|t}^j$ and $K_{k|t}^{i,j}$. The feedback gains enable the policy to adapt to NEV positions.

\subsection{Cost Function}

The cost function penalizes deviations from a reference trajectory and reference controls, while also promoting smooth control inputs. Formally, our objective function is
\begin{multline}
\label{eq:cost_fn}
  J_t(\mathbf{x}_t, \mathbf{u}_t) = \mathbb{E}_{j \sim p_t}\Bigg[\sum_{k=0}^{N-1} \left(\Delta x_{k+1|t}^j\right)^\top Q \Delta x_{k+1|t}^j \\
  + \left(\Delta u_{k|t}^{j}\right)^\top R_1 \Delta u_{k|t}^j \\
  + \left(u_{k+1|t}^j - u_{k|t}^j\right)^\top R_2 \left(u_{k+1|t}^j - u_{k|t}^j\right) \Bigg]
\end{multline}
where $\Delta x_{k+1|t}^j = x_{k+1|t}^j - x_{k+1|t}^\text{ref}$ is the deviation from the reference trajectory, and $\Delta u_{k|t}^j = u_{k|t}^j - u_{k|t}^\text{ref}$ is the deviation from the reference control. The weight matrices $Q \succ 0$, $R_1 \succ 0$, and $R_2 \succ 0$ control the relative importance of trajectory tracking, control tracking, and control smoothness, respectively. The reference controls $u_{k|t}^\text{ref}$ are derived from the reference trajectory as finite differences: $u_{k|t}^\text{ref} = (x_{k+1|t}^\text{ref} - x_{k|t}^\text{ref}) / \delta_t$, representing the velocities required to follow the reference path. The third term penalizes changes in control inputs between consecutive timesteps, promoting smooth and comfortable motion.

\section{EXPERIMENTS}\label{sec:expts}

\begin{table*}[ht!]
  \centering
  \caption{Prediction metrics across short-horizon (3s) and long-horizon (8s) evaluations. On PUP (with class uncertainty), HAICU outperforms Trajectron++ on most metrics. On nuPlan (without class uncertainty), HAICU cannot leverage its key advantage and the constant velocity baseline achieves competitive minADE/minFDE on long horizons. Values reported as mean $\pm$ standard error; ECE is computed as a single aggregate metric over all predictions. Entropy is not reported for the constant velocity baseline as it uses a fixed covariance matrix regardless of horizon. Bold indicates best.}
  \label{table:prediction_metrics}
  
  \vspace{0.5em}
  \textbf{(a) PUP Dataset}
  \vspace{0.3em}
  
  \begin{tabular}{c ccc ccc}
    \toprule
    & \multicolumn{3}{c}{3s Horizon} & \multicolumn{3}{c}{8s Horizon} \\
    \cmidrule(lr){2-4} \cmidrule(lr){5-7}
    Metric & T++ & HAICU & Const. & T++ & HAICU & Const. \\  
    \midrule
    minADE$_5$ (\unit{\m}) $[\downarrow]$ & $0.72\pm0.01$ & $\boldsymbol{0.63\pm0.01}$ & $1.29\pm0.02$ & $5.62\pm0.39$ & $\boldsymbol{5.00\pm0.38}$ & $5.17\pm0.37$ \\
    minFDE$_5$ (\unit{\m}) $[\downarrow]$ & $1.54\pm0.03$ & $\boldsymbol{1.36\pm0.02}$ & $2.79\pm0.04$ & $12.25\pm0.89$ & $\boldsymbol{11.75\pm0.92}$ & $12.24\pm0.89$ \\
    Avg NLL $[\downarrow]$ & $-1.14\pm0.03$ & $\boldsymbol{-1.48\pm0.03}$ & $199.07\pm6.57$ & $3.20\pm0.26$ & $\boldsymbol{1.72\pm0.22}$ & $3155\pm373$ \\
    Avg Entropy $[\downarrow]$ & $-0.39\pm0.01$ & $\boldsymbol{-0.68\pm0.02}$ & -- & $4.26\pm0.10$ & $\boldsymbol{2.55\pm0.09}$ & -- \\
    ECE $[\downarrow]$ & $0.16$ & $0.20$ & $\boldsymbol{0.14}$ & $0.16$ & $\boldsymbol{0.14}$ & $0.18$ \\
    \bottomrule
  \end{tabular}
  
  \vspace{1.5em}
  \textbf{(b) nuPlan Dataset}
  \vspace{0.3em}
  
  \begin{tabular}{c ccc ccc}
    \toprule
    & \multicolumn{3}{c}{3s Horizon} & \multicolumn{3}{c}{8s Horizon} \\
    \cmidrule(lr){2-4} \cmidrule(lr){5-7}
    Metric & T++ & HAICU & Const. & T++ & HAICU & Const. \\  
    \midrule
    minADE$_5$ (\unit{\m}) $[\downarrow]$ & $\boldsymbol{0.33\pm0.01}$ & $0.38\pm0.01$ & $0.57\pm0.01$ & $4.71\pm0.09$ & $5.70\pm0.12$ & $\boldsymbol{3.87\pm0.08}$ \\
    minFDE$_5$ (\unit{\m}) $[\downarrow]$ & $\boldsymbol{0.81\pm0.02}$ & $0.91\pm0.03$ & $1.35\pm0.03$ & $10.22\pm0.20$ & $12.44\pm0.26$ & $\boldsymbol{9.07\pm0.18}$ \\
    Avg NLL $[\downarrow]$ & $-1.50\pm0.03$ & $\boldsymbol{-1.68\pm0.04}$ & $40.15\pm1.68$ & $3.29\pm0.07$ & $\boldsymbol{1.95\pm0.05}$ & $1720\pm52$ \\
    Avg Entropy $[\downarrow]$ & $-0.38\pm0.02$ & $\boldsymbol{-0.73\pm0.02}$ & -- & $4.61\pm0.03$ & $\boldsymbol{2.74\pm0.03}$ & -- \\
    ECE $[\downarrow]$ & $0.17$ & $0.19$ & $\boldsymbol{0.08}$ & $\boldsymbol{0.13}$ & $\boldsymbol{0.13}$ & $0.14$ \\
    \bottomrule
  \end{tabular}
\end{table*}


\begin{table*}[ht!]
  \centering
  \caption{Closed-Loop Planning with reactive agents across short-horizon (3s) and long-horizon (8s) evaluations. On short horizons, T++ achieves the best progress, while HAICU leads in TTC and Closed Loop Score. On long horizons, HAICU UA-SMPC surpasses all other methods in closed-loop score, while T++ UA-SMPC shows a marked decline in performance. Values reported as mean $\pm$ standard error. Bold indicates best.}
  \label{table:closed_loop_metrics}
  
  \begin{tabular}{c ccc ccc}
    \toprule
    & \multicolumn{3}{c}{Short-Horizon (3s)} & \multicolumn{3}{c}{Long-Horizon (8s)} \\
    \cmidrule(lr){2-4} \cmidrule(lr){5-7}
    Metric & T++ & HAICU & Const. & T++ & HAICU & Const. \\  
    \midrule
    Ego Progress $[\uparrow]$ & $\boldsymbol{0.85\pm0.09}$ & $0.73\pm0.08$ & $0.78\pm0.08$ & $0.43\pm0.11$ & $0.68\pm0.08$ & $\boldsymbol{0.77\pm0.07}$ \\
    Jerk (\si{\meter\per\second^3}) $[\downarrow]$ & $0.16\pm0.06$ & $\boldsymbol{0.15\pm0.03}$ & $\boldsymbol{0.15\pm0.05}$ & $0.48\pm0.15$ & $0.18\pm0.04$ & $\boldsymbol{0.17\pm0.03}$ \\
    TTC (\unit{\s}) $[\uparrow]$ & $0.89\pm0.63$ & $\boldsymbol{1.44\pm0.57}$ & $0.71\pm0.36$ & $1.19\pm0.40$ & $\boldsymbol{2.52\pm0.65}$ & $1.69\pm0.75$ \\
    \midrule
    Closed Loop Score $[\uparrow]$ & $0.64\pm0.04$ & $\boldsymbol{0.72\pm0.05}$ & $0.70\pm0.04$ & $0.55\pm0.06$ & $\boldsymbol{0.79\pm0.04}$ & $0.72\pm0.04$ \\
    \bottomrule
  \end{tabular}
\end{table*}

To evaluate the impact of upstream uncertainty propagation on motion planning, we compare two different prediction-planning methods on the complete set of nuPlan challenge scenarios~\cite{caesar2021nuplan}: Trajectron++~\cite{SalzmannIvanovicEtAl2020} and HAICU~\cite{IvanovicLeeEtAl2022}, both paired with our proposed UA-SMPC. The two prediction architectures are highly similar, with the only difference being HAICU's incorporation of input class uncertainty. This distinction allows us to quantify the impact of uncertainty propagation and calibration on motion planning. We additionally introduce a simple prediction baseline, using a constant velocity model for both EV and NEV predictions, to serve as a benchmark for our experiments. The constant velocity baseline generates predictions by linearly extrapolating each agent's current velocity over the prediction horizon, assuming constant heading and a fixed covariance of $\sigma^2 = 0.02$ (selected after a brief hyperparameter sweep) for all future timesteps; its entropy is therefore constant at $\approx -1.07$. We evaluate prediction metrics across both short-horizon (3 seconds at 300ms intervals) and long-horizon (8 seconds at 800ms intervals) settings to assess model performance under varying planning horizons and prediction frequencies. 

\subsection{Methodology}
For each scenario, consisting of over 150 timesteps, both planning-prediction methods are provided a mission goal and roadblock information with the desired route path. Following the method of~\cite{huang2023gameformer}, a reference route is generated at each timestep using the roadblock information. This reference route includes 1,000 waypoints spaced 0.1 \unit{\m} apart and stretches 100 \unit{\m} ahead of the EV. To obtain the EV reference trajectory over the next N timesteps, the EV prediction from $f_k^\text{EV}$ is projected onto the reference route to maintain proper lane alignment. This reference trajectory, along with EV and NEV current states and histories, are passed into our UA-SMPC framework for path planning optimization. Evaluation is conducted in closed-loop manner, where each approach is compared on a per-timestep basis, with the initial states for all agents coming from the dataset.

\subsection{Metrics}
The two methods are compared on various common deterministic and probabilistic prediction and planning metrics. For prediction metrics, we are interested in comparing the fidelity of the output distributions, thus we measure:
\begin{itemize}
\item \textit{$\text{minADE}_5$}: ADE/FDE between the ground truth and best of 5 predicted output samples~\cite{GuptaJohnsonEtAl2018},
\item \textit{Average Negative Log-Likelihood (NLL)}: The average NLL of the ground truth trajectory under the predicted future distribution, and
\item \textit{Average Entropy}: The entropy of each GMM component distribution averaged across all agents and time.
 \item \textit{Expected Calibration Error (ECE)}: A measure of how well the predicted uncertainty estimates align with observed error frequencies~\cite{ovadia2019can}. Following~\cite{IvanovicHarrison2023}, we compute the average orthogonal distance between the empirical cumulative distribution of prediction errors and the ideal calibration curve, where well-calibrated predictions yield ECE values near zero.
\end{itemize}


To evaluate closed-loop planning, we use three metrics that quantify progress, comfort, and safety:
\begin{itemize}
    \item \textit{Ego Progress along Expert Route (Progress):} Ratio of overall ego progress to overall expert progress~\cite{caesar2021nuplan},
    \item \textit{Jerk (Comfort):} Rate of change of acceleration, which affects smoothness and passenger comfort~\cite{caesar2020nuscenes}, and
    \item \textit{Time to Collision (Safety):} Time required for EV to collide with another object assuming they continue to travel at a constant velocity~\cite{caesar2021nuplan}.
\end{itemize}

Additionally, we evaluate our methods using nuPlan's closed-loop scoring function as an overall comprehensive metric of planner performance.

\subsection{Prediction Results}
We train our Trajectron++ and HAICU prediction models on the Perceptual Uncertainty in Prediction (PUP) dataset, a real-world autonomous driving dataset that uniquely includes unfiltered class probabilities for each agent~\cite{IvanovicLeeEtAl2022}. Unlike most autonomous driving datasets (which lack class uncertainty data), the PUP dataset enables a direct comparison of uncertainty propagation's impact on motion planning across varying levels of uncertainty awareness.

\cref{table:prediction_metrics} summarizes the performance of Trajectron++, HAICU, and the constant velocity baseline evaluated with prediction metrics on the PUP and nuPlan datasets. On PUP, HAICU clearly outperforms Trajectron++ in both accuracy and calibration of the output distribution. In fact, our results are essentially identical to those in Table III of~\cite{IvanovicLeeEtAl2022}. On average, HAICU's predictions are more informative, achieving a lower NLL and lower entropy.

On nuPlan, where class uncertainty is not available, HAICU cannot leverage its key advantage over Trajectron++ (the propagation of class uncertainty) and its performance degrades relative to PUP. On short horizons, both learned models clearly outperform the constant velocity baseline in point accuracy. However, on 8-second horizons, both deteriorate significantly: HAICU is outperformed by the constant velocity baseline on minADE and minFDE, and Trajectron++ approaches it. This suggests that nuPlan scenarios exhibit more constant-velocity-like behavior than the PUP dataset. Despite this, both learned models maintain significantly better NLL than the constant velocity baseline, which suffers from catastrophically miscalibrated uncertainty estimates due to its fixed covariance.

Overall, prediction performance degrades noticeably as the horizon increases: NLL rises significantly from 3s to 8s for both learned models, and their point accuracy advantage over the constant velocity baseline largely disappears on long horizons. ECE, however, remains relatively stable ($\sim$0.13--0.20) across horizons for all models, indicating that the learned models are systematically underconfident from the start. Both HAICU and Trajectron++ exhibit higher ECE than the constant velocity baseline, particularly on short horizons. As we discuss further in~\cref{sec:sensitivity_analysis}, this is a consequence of their over-conservative predictions.
 

\subsection{Closed-Loop Quantitative Results} 
\cref{table:closed_loop_metrics} summarizes the performance of our constant velocity, Trajectron++, and HAICU-backed UA-SMPC in closed-loop nuPlan simulation with reactive agents, evaluated over both short-horizon (\SI{3}{\second}) and long-horizon (\SI{8}{\second}) planning. 

On short horizons, T++ achieves the best \textit{Ego Progress}, while HAICU leads in \textit{TTC} and \textit{Closed Loop Score}. The constant velocity baseline performs competitively, tying for the lowest jerk.

On long horizons, HAICU-backed UA-SMPC clearly surpasses all other methods in \textit{Closed Loop Score}. Furthermore, Trajectron++-backed UA-SMPC exhibits a marked decline in performance across all metrics. Notably, the constant velocity baseline achieves the best \textit{Ego Progress} on long horizons, while HAICU maintains the best \textit{Time to Collision}, indicating safer planning behavior.

These results can be connected to the prediction metrics in~\cref{table:prediction_metrics}. Despite HAICU's worse point accuracy on nuPlan (where class uncertainty is unavailable), its superior NLL/Entropy indicates better-calibrated distributions, which proves more valuable for closed-loop planning with reactive agents. We also observe that methods with higher NLL (i.e., more miscalibrated predictions) tend to achieve lower TTC, suggesting that poorly calibrated uncertainty estimates result in reduced safety margins. Furthermore, the constant velocity baseline's competitive performance aligns with its competitive minADE/minFDE on nuPlan, further supporting the observation in~\cref{table:prediction_metrics} that many nuPlan scenarios feature near-constant-velocity behavior.

\subsection{Qualitative Results} 
\begin{figure*}[h!]
  \centering
  \setlength{\tabcolsep}{0pt} 
  \renewcommand{\arraystretch}{0} 
  \begin{tabular}{ccc}
    \includegraphics[width=0.33\linewidth]{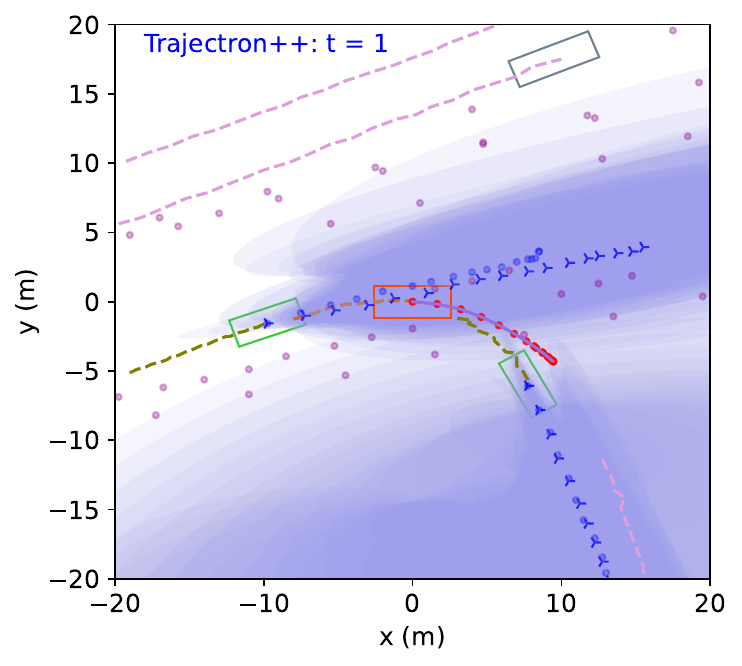} &
    \includegraphics[width=0.33\linewidth]{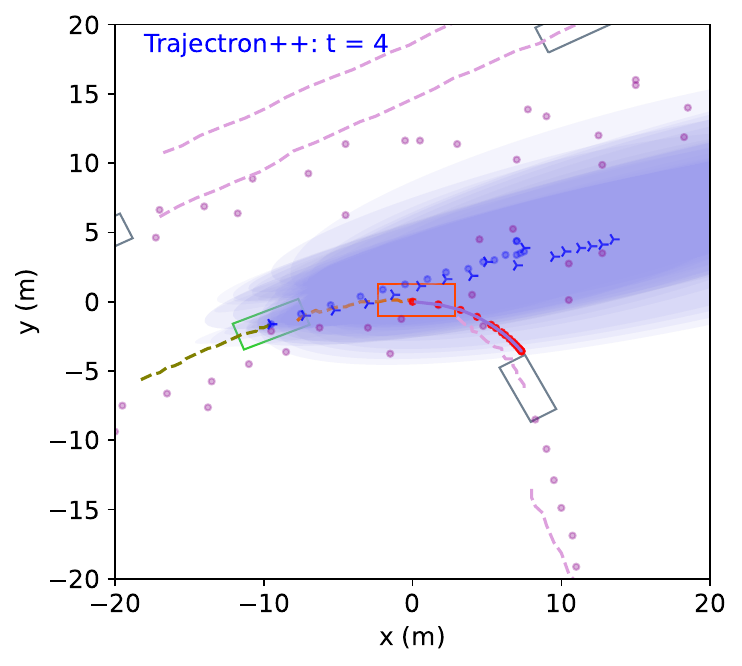} &
    \includegraphics[width=0.33\linewidth]{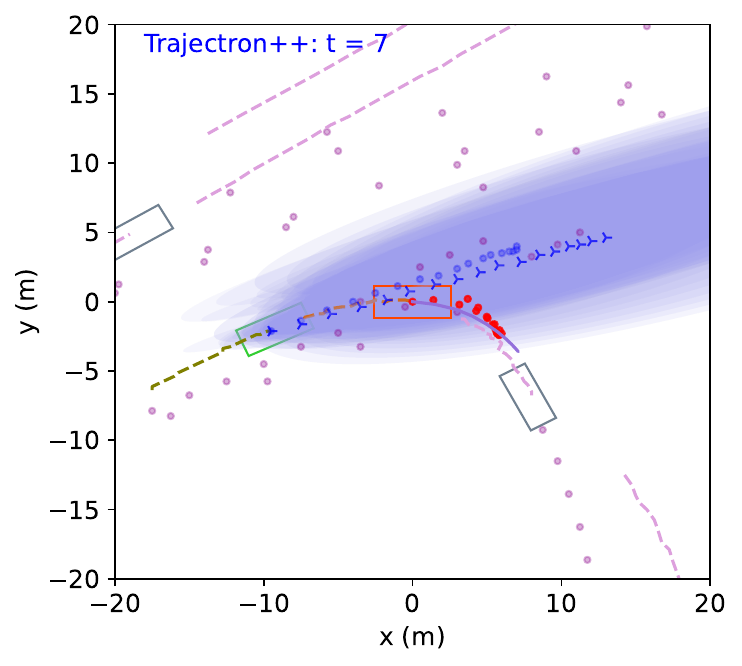} \\
    \includegraphics[width=0.33\linewidth]{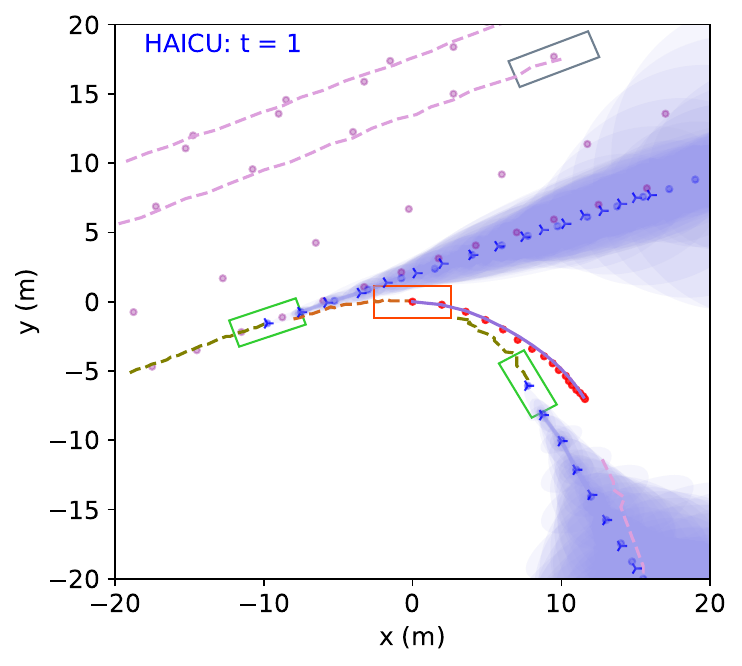} &
    \includegraphics[width=0.33\linewidth]{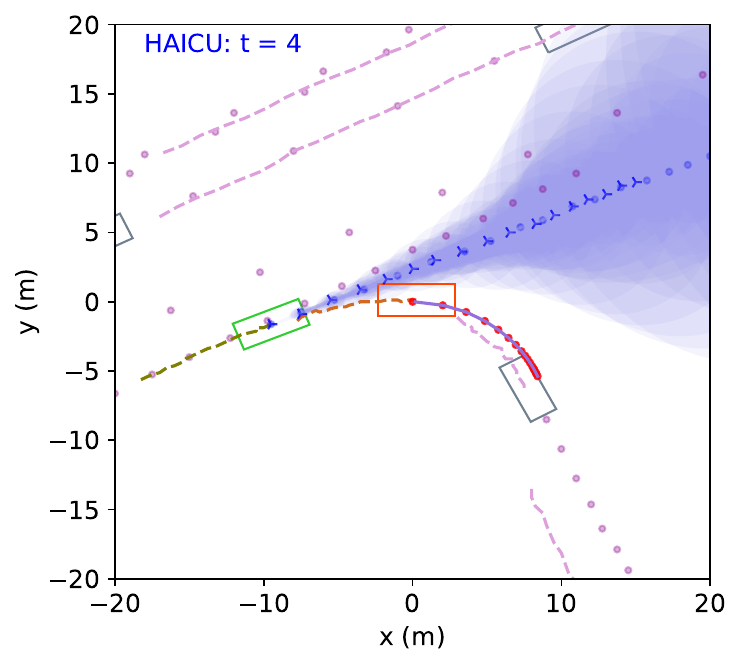} &
    \includegraphics[width=0.33\linewidth]{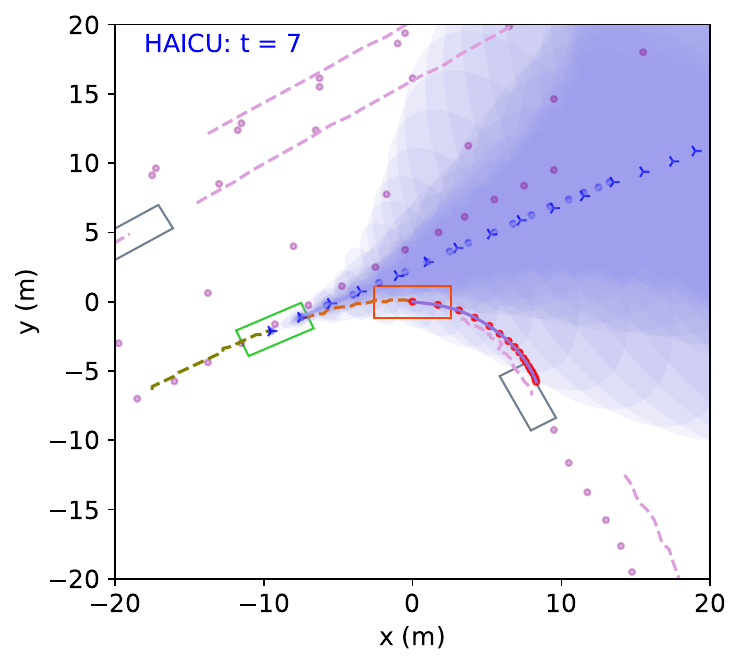} \\
  \end{tabular}
  \caption{A rollout comparison of UA-SMPC paired with Trajectron++~\cite{SalzmannIvanovicEtAl2020} (top) and HAICU~\cite{IvanovicLeeEtAl2022} (bottom) over three sequential timesteps in an open-loop right turn scenario. Trajectron++ fails to predict the trailing agent’s right turn, while HAICU accurately forecasts it with increasing likelihood, leading to a safer motion plan. See~\cref{fig:teaser} for legend.}
  \label{fig:qualitative}
\end{figure*}

Visually, HAICU produces more calibrated outputs than Trajectron++, capturing future trajectories with greater likelihood. In~\cref{fig:qualitative}, Trajectron++ consistently produces poorly calibrated predictions for nearby agents at each sampled timestep and fails to anticipate the potential right turn for the trailing nearby agent. In contrast, HAICU's predictions are well-calibrated to the scene, clearly accounting for the trailing agent's potential right turn and increasing its likelihood as the agent approaches the intersection.

Due to the calibration differences between prediction models, UA-SMPC converges to different control actions depending on the model used. With Trajectron++-backed UA-SMPC, the risk of a collision with the potentially right-turning agent fails to propagate, leading to a jerky motion plan that overlooks this risk at $t=7$. In contrast, the HAICU-backed UA-SMPC recognizes this collision risk and actively adjusts its motion plan to stay within the low probability regions of the predicted agent trajectory.

\subsection{Discussion}
\label{sec:discussion}
Our results show that HAICU-backed UA-SMPC demonstrates better generalization to complex driving scenarios, particularly in long-horizon closed-loop planning.

These results can be attributed to the difference in uncertainty awareness between Trajectron++ and HAICU. Since Trajectron++ lacks uncertainty propagation in its architecture, it tends to produce poorly calibrated predictions that fail to capture the true collision risks in a scene. As shown in~\cref{fig:qualitative}, T++ misses plausible future modes (e.g., a trailing agent's right turn), so its avoidance regions are centered on the wrong locations, leaving dangerous directions unprotected. In contrast, HAICU produces well-calibrated predictions that account for multiple modes of future action, leading to safer and smoother motion plans.

Quantitatively, this is reflected in \cref{table:closed_loop_metrics}: T++-backed UA-SMPC achieves higher \textit{Ego Progress} on short horizons not because it plans aggressively through known risks, but because its miscalibrated predictions fail to identify them, resulting in lower \textit{TTC} and higher \textit{Jerk}. On long horizons, HAICU-backed UA-SMPC achieves the best \textit{Closed Loop Score}, demonstrating the value of well-calibrated uncertainty propagation.

Importantly, our results show that for GMM-based uncertainty-aware planners such as UA-SMPC, distributional quality (NLL/Entropy) is a stronger predictor of planning performance than point accuracy (minADE/minFDE). Despite HAICU's worse minADE/minFDE on nuPlan (\cref{table:prediction_metrics}b), its superior NLL and entropy yield the best closed-loop score. This is expected, as our SMPC formulation plans directly over the predicted covariances, not just the means.

\subsection{Sensitivity Analysis}
\label{sec:sensitivity_analysis}

To investigate the sensitivity of our planning framework to uncertainty calibration, we conduct a sensitivity analysis by scaling the predicted covariance matrices $\Sigma_{k|t}^{i,j}$ from the GMM output (as defined in~\cref{sec:method}) by a factor $\alpha$, transforming them as $\Sigma_{k|t}^{i,j} \rightarrow \alpha \Sigma_{k|t}^{i,j}$ before passing them into the UA-SMPC planner. We evaluate the impact of this scaling on closed-loop planning performance with reactive agents across scaling factors $\alpha \in \{1/4, 1/3, 1/2, 1, 2, 3, 4, 5\}$, where $\alpha=1$ corresponds to the default (unscaled) predictions.

\begin{table*}[ht!]
  \centering
  \caption{HAICU sensitivity analysis: Closed-loop planning performance across covariance scaling factors $\alpha$.}
  \label{table:sensitivity_haicu}
  \footnotesize
  \begin{tabular}{c cccc cccc}
    \toprule
    & \multicolumn{4}{c}{Reduced Uncertainty ($\alpha \leq 1$)} & \multicolumn{4}{c}{Increased Uncertainty ($\alpha \geq 2$)} \\
    \cmidrule(lr){2-5} \cmidrule(lr){6-9}
    Metric & $1/4$ & $1/3$ & $1/2$ & $1$ & $2$ & $3$ & $4$ & $5$ \\
    \midrule
    Progress $[\uparrow]$ & $\boldsymbol{0.70\pm0.10}$ & $0.68\pm0.08$ & $0.62\pm0.09$ & $0.60\pm0.09$ & $0.61\pm0.11$ & $0.57\pm0.11$ & $0.60\pm0.11$ & $0.58\pm0.11$ \\
    Jerk $[\downarrow]$ & $\boldsymbol{0.17\pm0.04}$ & $0.18\pm0.04$ & $0.20\pm0.05$ & $0.28\pm0.05$ & $0.26\pm0.08$ & $0.26\pm0.06$ & $0.20\pm0.04$ & $0.18\pm0.04$ \\
    TTC $[\uparrow]$ & $2.27\pm0.68$ & $2.35\pm0.68$ & $2.27\pm0.86$ & $\boldsymbol{3.19\pm0.84}$ & $1.83\pm0.72$ & $1.83\pm0.72$ & $2.37\pm0.78$ & $2.04\pm0.69$ \\
    \midrule
    CL Score $[\uparrow]$ & $0.76\pm0.04$ & $\boldsymbol{0.77\pm0.04}$ & $0.71\pm0.04$ & $0.73\pm0.03$ & $0.71\pm0.05$ & $0.64\pm0.05$ & $0.66\pm0.05$ & $0.64\pm0.04$ \\
    \bottomrule
  \end{tabular}
\end{table*}

\begin{table*}[ht!]
  \centering
  \caption{T++ sensitivity analysis: Closed-loop planning performance across covariance scaling factors $\alpha$.}
  \label{table:sensitivity_tpp}
  \footnotesize
  \begin{tabular}{c cccc cccc}
    \toprule
    & \multicolumn{4}{c}{Reduced Uncertainty ($\alpha \leq 1$)} & \multicolumn{4}{c}{Increased Uncertainty ($\alpha \geq 2$)} \\
    \cmidrule(lr){2-5} \cmidrule(lr){6-9}
    Metric & $1/4$ & $1/3$ & $1/2$ & $1$ & $2$ & $3$ & $4$ & $5$ \\
    \midrule
    Progress $[\uparrow]$ & $\boldsymbol{0.51\pm0.13}$ & $0.41\pm0.11$ & $0.46\pm0.11$ & $0.47\pm0.11$ & $0.49\pm0.12$ & $0.44\pm0.11$ & $0.47\pm0.13$ & $0.49\pm0.12$ \\
    Jerk $[\downarrow]$ & $0.24\pm0.06$ & $0.24\pm0.11$ & $0.25\pm0.04$ & $0.24\pm0.05$ & $\boldsymbol{0.17\pm0.05}$ & $0.40\pm0.20$ & $0.24\pm0.04$ & $0.22\pm0.05$ \\
    TTC $[\uparrow]$ & $1.33\pm0.58$ & $\boldsymbol{2.18\pm0.85}$ & $2.05\pm0.83$ & $1.29\pm0.51$ & $1.54\pm0.88$ & $1.27\pm0.71$ & $1.41\pm0.95$ & $1.33\pm0.82$ \\
    \midrule
    CL Score $[\uparrow]$ & $\boldsymbol{0.58\pm0.06}$ & $0.58\pm0.05$ & $0.58\pm0.06$ & $0.56\pm0.06$ & $0.52\pm0.05$ & $0.51\pm0.06$ & $0.43\pm0.05$ & $0.47\pm0.04$ \\
    \bottomrule
  \end{tabular}
\end{table*}

The sensitivity analysis reveals several key findings. For HAICU (\cref{table:sensitivity_haicu}), reducing the covariance scaling factor improves overall performance: the best \textit{Closed Loop Score} (0.77) is achieved at $\alpha=1/3$, outperforming the default $\alpha=1$ (0.73). This suggests that HAICU's predictions are systematically over-conservative. We also observe a clear progress-safety tradeoff: as uncertainty decreases, \textit{Progress} increases (from 0.60 at $\alpha=1$ to 0.70 at $\alpha=1/4$) while \textit{TTC} decreases (from 3.19 to 2.27), as the planner becomes more aggressive with tighter uncertainty bounds. More specifically, when the region to avoid collision is tightened, the planner has more freedom to make progress (which our cost function encourages).

In contrast, T++ (\cref{table:sensitivity_tpp}) exhibits erratic behavior across scaling factors, with no clear monotonic trend in any metric. This instability reflects T++'s poor baseline calibration: since its uncertainty estimates are already miscalibrated, scaling them does not yield predictable improvements. Notably, even with optimal scaling ($\alpha \leq 1/2$), T++'s best \textit{Closed Loop Score} (0.58) remains significantly below HAICU's optimally-scaled performance (0.77).

\section{CONCLUSION}\label{sec:conclusion}
In this work, we present a novel UA-SMPC formulation with exact probabilistic collision avoidance and compare two prediction-planning pipelines with varying levels of upstream uncertainty propagation on nuPlan's closed-loop planning benchmark. Our results yield two key findings. First, for GMM-based uncertainty-aware planners, distributional quality (NLL/Entropy) is a stronger determinant of planning performance than point accuracy (minADE/minFDE). Despite HAICU's worse displacement errors on nuPlan, its superior distributional calibration leads to the best closed-loop score across both horizons. Second, our sensitivity analysis reveals that HAICU's well-calibrated predictions respond predictably to covariance scaling, whereas Trajectron++'s miscalibrated predictions exhibit erratic behavior, highlighting the importance of baseline calibration quality for robust planning.

{\bf Future Work.} To address the limitations of training our prediction models on the independent PUP dataset, an immediate next step is to train them directly on the nuPlan training set instead. This would, however, first require training and integrating a perception system to predict class uncertainty for each agent, as the nuPlan dataset does not natively provide this information. Another future direction is to introduce additional prediction-planning methods for more comprehensive comparison between different types of uncertainty propagation. For example, PSU-TF~\cite{IvanovicLinEtAl2022}, which shares similar architecture as Trajectron++ but includes state uncertainties, could be explored to evaluate the impact of state uncertainty propagation on perception-based motion planning.

{\bf Acknowledgements.} The Toyota Research Institute partially supported this work. This article solely reflects the opinions and conclusions of its authors and not TRI or any other Toyota entity.

\bibliographystyle{IEEEtran}
\bibliography{main}

\addtolength{\textheight}{-10.5cm}   




\end{document}